\documentclass[10pt,twocolumn,letterpaper]{article}

\usepackage{cvpr}
\usepackage{times}
\usepackage{epsfig}
\usepackage{graphicx}
\usepackage{amsmath}
\usepackage{amssymb}

\usepackage{graphicx}
\usepackage{amsmath,amssymb} 
\usepackage{makecell}
\usepackage{color}
\usepackage{enumitem}
\usepackage{multirow}
\usepackage{booktabs}

\DeclareMathOperator*{\argmax}{arg\,max}

\usepackage[breaklinks=true,bookmarks=false]{hyperref}

\cvprfinalcopy 

\def\cvprPaperID{3566} 

\ifcvprfinal\fi
\begin{document}

\title{
	Streamlined Dense Video Captioning
} 

\author{
    Jonghwan Mun$^{1,5}\thanks{This work was done during the internship program at Snap Research.}$ \hspace{0.5cm}
    Linjie Yang$^{2}$ \hspace{0.5cm}
    Zhou Ren$^{3}$ \hspace{0.5cm}
    Ning Xu$^{4}$ \hspace{0.5cm}
    Bohyung Han$^{5}$ \\
	\hspace{-0.3cm}
	$^{1}$POSTECH \hspace{0.05cm}
	$^{2}$ByteDance AI Lab \hspace{0.05cm}
	$^{3}$Wormpex AI Research \hspace{0.05cm}
	$^{4}$Amazon Go \hspace{0.05cm}
	$^{5}$Seoul National University \\
	$^{1}${\small\texttt{jonghwan.mun@postech.ac.kr}} \hspace{0.05cm}
	$^{2}${\small\texttt{linjie.yang@bytedance.com}} \hspace{0.05cm}
	$^{3}${\small\texttt{zhou.ren@bianlifeng.com}}\hspace{0.05cm} \\
	$^{4}${\small\texttt{ninxu@amazon.com}} \hspace{0.05cm}
	$^{5}${\small\texttt{bhhan@snu.ac.kr}}\\
}

\maketitle

\begin{abstract}
Dense video captioning is an extremely challenging task since accurate and coherent description of events in a video requires holistic understanding of video contents as well as contextual reasoning of individual events.
Most existing approaches handle this problem by first detecting event proposals from a video and then captioning on a subset of the proposals.
As a result, the generated sentences are prone to be redundant or inconsistent since they fail to consider temporal dependency between events.
To tackle this challenge, we propose a novel dense video captioning framework, which models temporal dependency across events in a video explicitly and leverages visual and linguistic context from prior events for coherent storytelling.
This objective is achieved by 1) integrating an event sequence generation network to select a sequence of event proposals adaptively, and 2) feeding the sequence of event proposals to our sequential video captioning network, which is trained by reinforcement learning with two-level rewards---at both event and episode levels---for better context modeling.
The proposed technique achieves outstanding performances on ActivityNet Captions dataset in most metrics.
\end{abstract}

\section{Introduction}
\label{sec:introduction}

Understanding video contents is an important topic in computer vision.
Through the introduction of large-scale datasets~\cite{tgif-cvpr2016, xu2016msr} and the recent advances of deep learning technology, research towards video content understanding is no longer limited to activity classification or detection and addresses more complex tasks including video caption generation~\cite{baraldi2017hierarchical, gan2017semantic, pan2016hierarchical, pan2016jointly, pan2017video, venugopalan2015sequence, venugopalan2015translating, wang2018reconstruction, wang2018m3, xiong2018move, yao2015describing, yu2016video, yu2017end}.

\begin{figure}[t]
	\centering
	\includegraphics[width=1.00\linewidth]{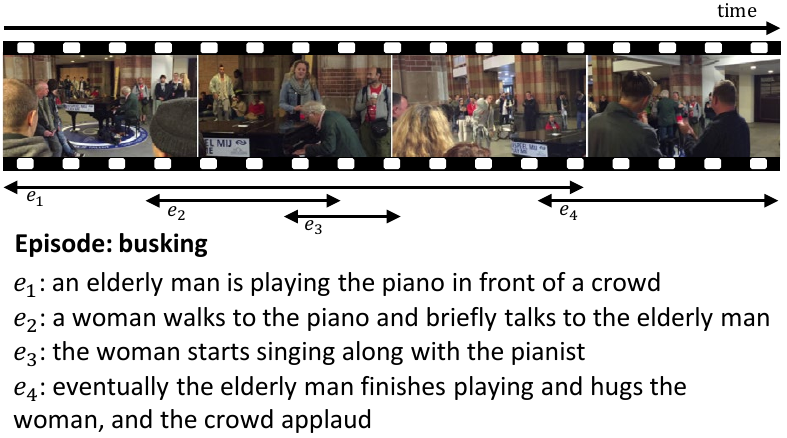}
	\caption{
		An example of dense video captioning about a {\it busking} episode, which is composed of four interdependent events.
	}
	\label{fig:example}
	\vspace{-0.2cm}
\end{figure}

Video  captions are effective for holistic video description.
However, since videos usually contain multiple interdependent events in context of a video-level story (\ie episode), a single sentence may not be sufficient to describe videos.
Consequently, dense video captioning task~\cite{krishna2017dense} has been introduced and getting more popular recently.
This task is conceptually more complex than simple video captioning since it requires detecting individual events in a video and understanding their context.
Fig.~\ref{fig:example} presents an example of dense video captioning for a {\it busking} episode, which is composed of four ordered events.
%
%
Despite the complexity of the problem, most existing methods~\cite{krishna2017dense, li2018jointly, wang2018bidirectional, zhou2018end} are limited to describing an event using two subtasks---event detection and event description---in which an event proposal network is in charge of detecting events and a captioning network generates captions for the selected proposals independently.

\begin{figure}[!t]
	\centering
	\includegraphics[width=0.98\linewidth]{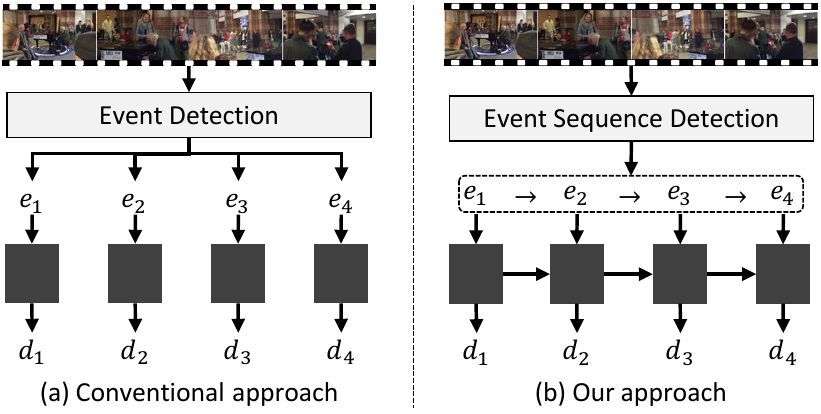}
	\caption{
		Comparison between the existing approaches and ours for dense video captioning. 
		Our algorithm generates captions for events sequentially conditioned on the prior ones by detecting an event sequence in a video.
	}
	\label{fig:problem_definition}
	\vspace{-0.2cm}
\end{figure}

We propose a novel framework for dense video captioning, which considers the temporal dependency of the events.
Contrary to existing approaches shown in Fig.~\ref{fig:problem_definition}(a), our algorithm detects event sequences from videos and generates captions sequentially, where each caption is conditioned on prior events and captions as illustrated in Fig.~\ref{fig:problem_definition}(b).
Our algorithm has the following procedure. 
First, given a video, we obtain a set of candidate event proposals from an event proposal network.
Then, an event sequence generation network selects a series of ordered events adaptively from the event proposal candidates.
Finally, we generate captions for the selected event proposals using a sequential captioning network. 
The captioning network is trained via reinforcement learning using both event and episode-level rewards; the event-level reward allows to capture specific content in each event precisely while the episode-level reward drives all generated captions to make a coherent story.

The main contributions of the proposed approach are summarized as follows:
\begin{itemize}[label=$\bullet$]
	\item
	We propose a novel framework of detecting event sequences for dense video captioning. The proposed event sequence generation network allows the captioning network to model temporal dependency between events and generate a set of coherent captions to describe an episode in a video. 
	\item 
	We present reinforcement learning with two-level rewards, \emph{episode} and \emph{event} levels, which drives the captioning model to boost coherence across generated captions and quality of description for each event.
	\item
	The proposed algorithm achieves the state-of-the-art performance on the ActivityNet Captions dataset with large margins compared to the methods based on the existing framework.
\end{itemize}

The rest of the paper is organized as follows.
We first discuss related works for our work in Section~\ref{sec:related}.
The proposed method and its training scheme are described in Section~\ref{sec:framework} and \ref{sec:training} in detail, respectively.
We present experimental results in Section~\ref{sec:experiments}, and conclude this paper in Section~\ref{sec:conclusion}.

\section{Related Work}
\label{sec:related}

\subsection{Video Captioning}
\label{sec:video_captioning}

Recent video captioning techniques often incorporate the encoder-decoder framework inspired by success in image captioning~\cite{mun2017text, ren2017deep, rennie2017self, vinyals2015show, xu2015show}.
Basic algorithms~\cite{venugopalan2015sequence,venugopalan2015translating} encode a video using Convolutional Neural Networks (CNNs) or Recurrent Neural Networks (RNNs), and decode the representation into a natural sentence using RNNs.
Then various techniques are proposed to enhance the quality of generated captions by integrating temporal attention~\cite{yao2015describing}, joint embedding space of sentences and videos~\cite{pan2016jointly}, hierarchical recurrent encoder~\cite{baraldi2017hierarchical, pan2016hierarchical}, attribute-augmented decoder~\cite{gan2017semantic, pan2017video, yu2017end}, multimodal memory~\cite{wang2018m3}, and reconstruction loss~\cite{wang2018reconstruction}.
Despite their impressive performances, they are limited to describing a video using a single sentence and can be applied only to a short video containing a single event.
Thus, Yu~\etal~\cite{yu2016video} propose a hierarchical recurrent neural network to generate a paragraph for a long video, while Xiong~\etal~\cite{xiong2018move} introduce a paragraph generation method based on event proposals, where an event selection module determines which proposals need to be utilized for caption generation in a progressive way.
Contrary to these tasks, which simply generate a sentence or paragraph for an input video, dense video captioning requires localizing and describing events at the same time.

\subsection{Dense Video Captioning}
\label{sec:densevidcap}

Recent dense video captioning techniques typically attempt to solve the problem using two subtasks---event detection and caption generation~\cite{krishna2017dense, li2018jointly, wang2018bidirectional, zhou2018end}; an event proposal network finds a set of candidate proposals and a captioning network is employed to generate a caption for each proposal independently.
The performance of the methods is affected by the manual thresholding strategies to select the final event proposals for caption generation.

Based on the framework, Krishna~\etal~\cite{krishna2017dense} adopt a multi-scale action proposal network~\cite{escorcia2016daps}, and introduce a captioning network that exploits visual context from past and future events with an attention mechanism.
In \cite{wang2018bidirectional}, a bidirectional RNN is employed to improve the quality of event proposals and a context gating mechanism in caption generation is proposed to adaptively control the contribution of surrounding events.
Li~\etal~\cite{li2018jointly} incorporate temporal coordinate and descriptiveness regressions for precise localization of event proposals, and adopt the attribute-augmented captioning network~\cite{yao2017boosting}.
Rennie~\etal~\cite{zhou2018end} utilize a self-attention~\cite{vaswani2017attention} for event proposal and captioning networks, and propose a masking network for conversion of the event proposals to differentiable masks and end-to-end learning of the two networks.

In contrast to the prior works, our algorithm identifies a small set of representative event proposals (\ie, event sequences) for sequential caption generation, which enables us to generate coherent and comprehensive captions by exploiting both visual and linguistic context across selected events. 
Note that the existing works fail to take advantage of linguistic context since the captioning network is applied to event proposals independently.

\section{Our Framework}
\label{sec:framework}
This section describes our main idea and the deep neural network architecture for our algorithm in detail.

\begin{figure*}[t]
	\centering
	\includegraphics[width=0.97\linewidth]{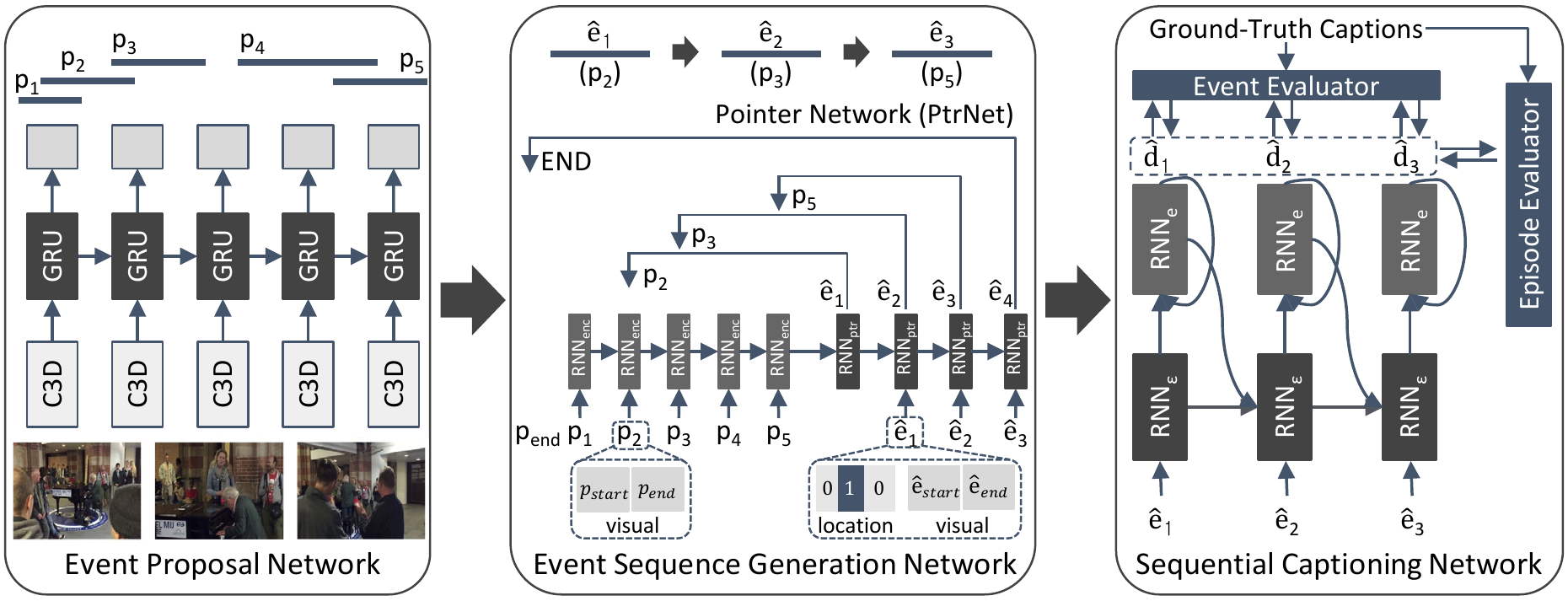}
	\caption{
		Overall framework of the proposed algorithm. 
		Given an input video, our algorithm first extracts a set of candidate event proposals ($p_1,p_2,p_3,p_4,p_5$) using the Event Proposal Network (Section~\ref{sec:proposalnet}).
		From the candidate set, the Event Sequence Generation Network detects an event sequence ($\hat{e}_1\rightarrow\hat{e}_2\rightarrow\hat{e}_3$) by selecting one out of the candidate event proposals (Section~\ref{sec:event_sequencenet}).
		Finally, the Sequential Captioning Network takes the detected event sequence and sequentially generates captions ($\hat{d}_1, \hat{d}_2, \hat{d}_3$) conditioned on preceding events (Section~\ref{sec:sequence_captioning}).
		The three models are trained in a supervised manner (Section~\ref{sec:supervised_leanring}) and then the Sequential Captioning Network is optimized additionally with reinforcement learning using two-level rewards (Section~\ref{sec:reinforcement_learning}).
	}
	\label{fig:architecture}
	\vspace{-0.2cm}
\end{figure*}

\subsection{Overview}
\label{sec:overview}

Let a video $V$ contain a set of events $\mathcal{E} = \{ e_1, \dots, e_N \}$ with corresponding descriptions $\mathcal{D} = \{ d_1, \dots, d_N \}$, where $N$ events are temporally localized using their starting and ending time stamps.
Existing methods~\cite{krishna2017dense, li2018jointly, wang2018bidirectional, zhou2018end} typically divide the whole problem into two steps: event detection followed by description of detected events.
These algorithms train models by minimizing the sum of negative log-likelihoods of event and caption pairs as follows:
\begin{align}
	\mathcal{L}&=\sum_{n=1}^{N} -\log p(d_n,e_n| {V}) \nonumber \\
	           &=\sum_{n=1}^{N} -\log p(e_n|V)p(d_n|e_n, {V}).
\label{eq:objective}
\end{align}

However, events in a video have temporal dependency and should be on a story about a single topic.
Therefore, it is critical to identify an ordered list of events to describe a coherent story corresponding to the episode, the composition of the events.
With this in consideration, we formulate dense video captioning as detection of an event sequence followed by sequential caption generation as follows:
\begin{align}
	\mathcal{L}&=-\log p(\mathcal{E}, \mathcal{D}| V) \nonumber \\
			   &=-\log p(\mathcal{E}| V)\prod_{n=1}^{N}p(d_n | d_{1},\dots,d_{n-1}, \mathcal{E}, V).
\label{eq:objective}
\end{align}

The overall framework of our proposed algorithm is illustrated in Fig.~\ref{fig:architecture}.
For a given video, a set of candidate event proposals is generated by the event proposal network.
Then, our event sequence generation network provides a series of events by selecting one of candidate event proposals sequentially, where the selected proposals correspond to events comprising an episode in the video.
Finally, we generate captions from the selected proposals using the proposed sequential captioning network, where each caption is generated conditioned on preceding proposals and their captions.
The captioning network is trained via reinforcement learning using event and episode-level rewards.

\subsection{Event Proposal Network (EPN)}
\label{sec:proposalnet}

EPN plays a key role in selecting event candidates.
We adopt Single-Stream Temporal action proposals (SST)~\cite{buch2017sst} due to its good performance and efficiency in finding semantically meaningful temporal regions via a single scan of videos.
SST divides an input video into a set of non-overlapping segments with a fixed length (\eg, 16 frames), where the representation of each segment is given by a 3D convolution (C3D) network~\cite{tran2015learning}.
By treating each segment as an ending point of an event proposal, SST identifies its matching starting points from the $k$ preceding segments, which are represented by $k$-dimensional output vector from a Gated Recurrent Unit (GRU) at each time step.
After extracting the top 1,000 event proposals, we obtain $M$ candidate proposals, $\mathcal{P} = \{p_1, \dots, p_M \}$, by eliminating highly overlapping ones using non-maximum suppression.
Note that EPN provides a representation of each proposal $p \in \mathcal{P}$, which is a concatenated vector of two hidden states at starting and ending segments in SST.
This visual representation, denoted by $\text{Vis}(p)$, is utilized for the other two networks.

\subsection{Event Sequence Generation Network (ESGN)} 
\label{sec:event_sequencenet}

Given a set of candidate event proposals, ESGN selects a series of events that are highly correlated and make up an episode for a video.
To this ends, we employ a Pointer Network (PtrNet)~\cite{vinyals2015pointer} that is designed to produce a distribution over the input set using a recurrent neural network by adopting an attention module.
PtrNet is well-suited for selecting an ordered subset of proposals and generating coherent captions with consideration of their temporal dependency.

As shown in Fig.~\ref{fig:architecture}, we first encode a set of candidate proposals, $\mathcal{P}$, by feeding proposals to an encoder RNN in an increasing order of their starting times, and initialize the first hidden state of PtrNet with the encoded representations to guide proposal selection.
At each time step in PtrNet, we compute likelihoods $a_t$ over the candidate event proposals and select a proposal with the highest likelihood out of all available proposals.
The procedure is repeated until PtrNet happens to select the \textit{END} event proposal, $p_{\text{end}}$, which is a special proposal to indicate the end of an event sequence.

The whole process is summarized as follows:
\begin{align}
h^{\text{ptr}}_0 &= \text{RNN}_{\text{enc}}({\text{Vis}(p_1), \dots, \text{Vis}(p_M)}), \\
h^{\text{ptr}}_t& = \text{RNN}_{\text{ptr}}(u(\hat{e}_{t-1}), h^{\text{ptr}}_{t-1}), \\
a_t& = \text{ATT}(h^{\text{ptr}}_t, u(p_{0}), \dots, u(p_{M})),
\end{align}
where $h^\text{ptr}$ is a hidden state in PtrNet, $\text{ATT}()$ is an attention function computing confidence scores over proposals, and the representation of proposal $p$ in PtrNet, $u(p)=[\text{Loc}(p);\text{Vis}(p)]$, is given by visual information $\text{Vis}(p)$ as well as the location information $\text{Loc}(p)$.
Also, $\hat{e}_{t}$ is a selected event proposal at time step $t$, which is given by
\begin{equation}
	\hat{e}_t =  p_{j^*} \text{  and~~} j^*=\argmax_{j\in\{0,\dots,M\}}~ a^j_t,
\end{equation}
where $p_0$ corresponds to $p_\text{end}$.
Note that the location feature, $\text{Loc}(p)$, is a binary mask vector, where the elements corresponding to temporal intervals of an event are set to 1s and 0s otherwise.
This is useful to identifying and disregarding proposals that overlap strongly with previous selections.

Our ESGN has clear benefits for dense video captioning.
Specifically, it determines the number and order of events adaptively, which facilitates compact, comprehensive and context-aware caption generation.
Noticeably, there are too many detected events in existing approaches (\eg, $\geq 50$) given by manual thresholding.
On the contrary, ESGN detects only 2.85 on average, which is comparable to the average number of events per video in ActivityNet Caption dataset, 3.65.
Although sorting event proposals is an ill-defined problem, due to their two time stamps (starting and ending points), ESGN naturally learns the number and order of proposals based on semantics and contexts in individual videos in a data-driven manner.

\subsection{Sequential Captioning Network (SCN)}
\label{sec:sequence_captioning}

SCN employs a hierarchical recurrent neural network to generate coherent captions based on the detected event sequence $\hat{\mathcal{E}}=\{\hat{e}_1, \dots, \hat{e}_{N_s}\}$, where $N_s (\leq M)$ is the number of selected events.
As shown in Fig.~\ref{fig:architecture}, SCN consists of two RNNs---an episode RNN and an event RNN---denoted by $\text{RNN}_{\mathcal{E}}$ and $\text{RNN}_{e}$, respectively.
The episode RNN takes the proposals in a detected event sequence one by one and models the state of an episode implicitly, while the event RNN generates words in caption sequentially for each event proposal conditioned on the implicit representation of the episode, \ie, based on the current context of the episode.

Formally, the caption generation process for the $t^\text{th}$ event proposal $\hat{e}_t$ in the detected event sequence is given by
\begin{align}
    r_t& = \text{RNN}_{\mathcal{E}}(\text{Vis}(\hat{e}_t), g_{t-1}, r_{t-1}), \\
    g_t & = \text{RNN}^*_{e}( \text{C3D}(\hat{e}_t), \text{Vis}(\hat{e}_t), r_t),
\end{align}
where $r_t$ is an episodic feature from the $t^\text{th}$ event proposal, and $g_t$  is a generated caption feature given by the last hidden state of the unrolled (denoted by $*$) event RNN.
$\text{C3D}(\hat{e}_t)$ denotes a set of C3D features for all segments lying in temporal intervals of $t^\text{th}$ event proposal.
The episode RNN provides the current episodic feature so that the event RNN generates context-aware captions, which are given back to the episode RNN.

Although both networks can be implemented with any RNNs conceptually, we adopt a single-layer Long Short-Term Memory (LSTM) with a 512 dimensional hidden state as the episode RNN, and a captioning network with temporal dynamic attention and context gating (TDA-CG) presented in~\cite{wang2018bidirectional} as the event RNN.
TDA-CG generates words from a feature computed by gating a visual feature $\text{Vis}(e)$ and an attended feature obtained from segment feature descriptors $\text{C3D}(e)$.

Note that sequential captioning generation scheme enables to exploit both visual context (\ie how other events look) and linguistic context (\ie how other events are described) across events, and allows us to generate captions in an explicit context.
Although existing methods~\cite{krishna2017dense,wang2018bidirectional} also utilize context for caption generation, they are limited to visual context and model with no linguistic dependency due to their architectural constraints from independent caption generation scheme, which would result in inconsistent and redundant caption generation.

\section{Training}
\label{sec:training}

We first learn the event proposal network and fix its parameters during training of the other two networks.
We train the event sequence generation network and the sequential captioning network in a supervised manner, and further optimize the captioning network based on reinforcement learning with two-level rewards---event and episode levels.

\subsection{Supervised Learning}
\label{sec:supervised_leanring}

\paragraph{Event Proposal Network}
Let $c^k_t$ be the confidence of the $k^\text{th}$ event proposal at time step $t$ in EPN, which is SST~\cite{buch2017sst} in our algorithm.
Denote the ground-truth label of the proposal by $y^k_t$, which is set to 1 if the event proposal has a temporal Intersection-over-Union (tIoU) with ground-truth events larger than 0.5, and 0 otherwise.
Then, for a given video $V$ and ground-truth labels $y$, we train EPN by minimizing a following weighted binary cross entropy loss:
\begin{align}
	\mathcal{L}_{\text{EPN}}&(V,\mathcal{Y}) = \nonumber \\
	& -\sum^{T_c}_{t=1}\sum^K_{k=1} y^k_t \log c^k_t + (1-y^k_t) \log(1-c^k_t), 
	\label{eq:sup_proposalnet}
\end{align}
where $\mathcal{Y} = \{y_t^k | 1 \leq t \leq T_c, 1 \leq k \leq K \}$, $K$ is the number of proposals containing each segment at the end and $T_c$ is the number of segments in the video.

\paragraph{Event Sequence Generation Network}
For a video with ground-truth event sequence $\mathcal{E}=\{e_1, \dots, e_N\}$ and a set of candidate event proposals $\mathcal{P}=\{p_1, \dots, p_M\}$,
the goal of ESGN is to select a proposal $p$ highly overlapping with the ground-truth event $e$, which is achieved by minimizing the following sum of binary cross entropy loss:
\begin{align}
	\mathcal{L}_{\text{ESGN}}(V, \mathcal{P}, \mathcal{E}) = &-\sum^{N}_{n=1}\sum^{M}_{m=1} 	\text{tIoU}(p_m, e_n)\log a^m_n  \\
	&+ (1-\text{tIoU}(p_m,e_n)) \log(1- a^m_n), \nonumber
\end{align}
where $\text{tIoU}(\cdot, \cdot)$ is a temporal Intersection-over-Union value between two proposals, and $a^m_n$ is the likelihood that the $m^{\text{th}}$ event proposal is selected as the $n^{\text{th}}$ event.

\paragraph{Sequential Captioning Network}
We utilize the ground-truth event sequence $\mathcal{E}$ and its descriptions $\mathcal{D}$ to learn our SCN via the \textit{teacher forcing} technique~\cite{williams1989learning}.
Specifically, to learn two RNNs in SCN, we provide episode RNN and event RNN with ground-truth events and captions as their inputs, respectively.
Then, the captioning network is trained by minimizing negative log-likelihood over words of the ground-truth captions as follows:
\begin{align}
	\mathcal{L}_{\text{SCN}}(V, \mathcal{E}, \mathcal{D})
	&= -\sum^{N}_{n=1} \log p(d_n|e_n)  \\
	&= -\sum^{N}_{n=1}\sum^{T_{d_n}}_{t=1} \log p(w^t_n|w^1_n, \dots, w^{t-1}_n, e_n), \nonumber
\end{align}
where $p(\cdot)$ denotes a predictive distribution over word vocabulary from the event RNN, and $w^t_n$ and $T_{d_n}$ mean the $t^{\text{th}}$ ground-truth word and the length of ground-truth description for the $n^{\text{th}}$ event.

\begin{table*}[!t]
	\centering
	\caption{
		Event detection performances including recall and precision at four thresholds of temporal intersection of unions (@tIoU) on the ActivityNet Captions validation set.
		The bold-faced numbers mean the best performance for each metric.
	}
	\vspace{0.1cm}
	\scalebox{0.9}{
		\begin{tabular}{c|ccccc|ccccc}
			\toprule
			\multirow{2}{*}{Method} & \multicolumn{5}{c|}{Recall (@tIoU)} & \multicolumn{5}{c}{Precision (@tIoU)} \\ 
			& @0.3 & @0.5 & @0.7 & @0.9 & Average & @0.3 & @0.5 & @0.7 & @0.9 & Average \\ 
			\hline\hline
			MFT~\cite{xiong2018move} & 46.18 & 29.76 & 15.54 & 5.77 & 24.31 & 86.34 & 68.79 & 38.30 & \textbf{12.19} & 51.41 \\
			ESGN (ours) & \textbf{93.41} & \textbf{76.40} & \textbf{42.40} & \textbf{10.10} & \textbf{55.58} & \textbf{96.71} & \textbf{77.73} & \textbf{44.84} & 10.99 & \textbf{57.57} \\
			\bottomrule
		\end{tabular}
	}
	\label{table:detection}
\end{table*}

\begin{table*}[!t]
	\centering
	\caption{
		Dense video captioning results including Bleu@N (B@N), CIDEr (C) and METEOR (M) for our model and other state-of-the-art methods on ActivityNet Captions validation set.
		We report performances obtained from both ground-truth (GT) proposals and learned proposals.
		Asterisk ($\ast$) stands for the methods re-evaluated using the newer evaluation tool and star ($\star$) indicates the methods exploiting additional modalities (\eg optical flow and attribute) for video representation.
		The bold-faced numbers mean the best for each metric.
	}
	\vspace{0.1cm}
	\scalebox{0.90}{
		\begin{tabular}{c|cccccc|cccccc}
			\toprule
			\multirow{2}{*}{Method} & \multicolumn{6}{c|}{with GT proposals} & \multicolumn{6}{c}{with learned proposals} \\ 
			& B@1 & B@2 & B@3 & B@4 & C & M & B@1 & B@2 & B@3 & B@4 & C & M \\
			\hline\hline 
			
			DCE~\cite{krishna2017dense} & 18.13 & 8.43 & 4.09 & 1.60 & 25.12 & 8.88 & 10.81 & 4.57 & 1.90 & 0.71 & 12.43 & 5.69 \\
			DVC~\cite{li2018jointly}{$^{\star}$} & 19.57 & 9.90 & 4.55 & 1.62 & 25.24 & 10.33 & 12.22 & 5.72 & 2.27 & 0.73 & 12.61 & 6.93  \\
			Masked Transformer~\cite{zhou2018end}{$^{\ast\star}$} & 23.93 & \textbf{12.16} & \textbf{5.76} & \textbf{2.71} & \textbf{47.71} & 11.16 & 9.96 & 4.81 & 2.42 & 1.15 & 9.25 & 4.98 \\
			TDA-CG~\cite{wang2018bidirectional}{$^{\ast}$} & - & - & - & - & - & 10.89 & 10.75 & 5.06 & 2.55 & \textbf{1.31} & 7.99 & 5.86 \\
			
			MFT~\cite{xiong2018move} & - & - & - & - & - & - & 13.31 & 6.13 & 2.82 & 1.24 & 21.00 & 7.08 \\
			SDVC (ours) & \textbf{28.02} & 12.05 & 4.41 & 1.28 & 43.48 & \textbf{13.07} & \textbf{17.92} & \textbf{7.99} & \textbf{2.94} & 0.93 & \textbf{30.68} & \textbf{8.82} \\
			\bottomrule
		\end{tabular}
	}
	\label{table:captioning}
	\vspace{-0.2cm}
\end{table*}

\subsection{Reinforcement Learning}
\label{sec:reinforcement_learning}

Inspired by the success in image captioning task~\cite{ren2017deep,rennie2017self}, we further employ reinforcement learning to optimize SCN.
While similar to the self-critical sequence training~\cite{rennie2017self} approach, the objective of learning our captioning network is revised to minimize the negative expected rewards for sampled captions.
The loss is formally given by
\begin{align}
	\mathcal{L}^{\text{RL}}_{\text{SCN}}(V, \hat{\mathcal{E}}, \mathcal{\hat{D}})
	= -\sum^{N_s}_{n=1}\mathbb{E}_{\hat{d}_n}\left[ \text{R}(\hat{d}_n)\right],
\end{align}
where $\mathcal{\hat{D}}=\{\hat{d}_1, \dots, \hat{d}_{N_S}\}$ is a set of sampled descriptions from the detected event sequence $\mathcal{\hat{E}}$ with $N_s$ events from ESGN, and $\text{R}(\hat{d})$ is a reward value for the individual sampled description $\hat{d}$.
Then, the expected gradient on the sample set $\mathcal{\hat{D}}$ is given by
\begin{align}
	\nabla\mathcal{L}^{\text{RL}}_{\text{SCN}}(V, \hat{\mathcal{E}}, \mathcal{\hat{D}}) 
	&= -\sum^{N_s}_{n=1} \mathbb{E}_{\hat{d}_n}\left[ \text{R}(\hat{d}_n) \nabla \log p(\hat{d}_n) \right] \nonumber \\
	&\approx -\sum^{N_s}_{n=1} \text{R}(\hat{d}_n) \nabla \log p(\hat{d}_n).
\end{align}

We adopt a reward function with two levels: episode and event levels. 
This encourages models to generate coherent captions by reflecting the overall context of videos, while facilitating the choices of better word candidates in describing individual events depending on the context.
Also, motivated by \cite{gu2017stack, ren2017deep, rennie2017self}, we use the rewards obtained from the captions generated with ground-truth proposals as baselines, which is helpful to reduce the variance of the gradient estimate.
This drives models to generate captions at least as competitive as the ones generated from ground-truth proposals, although the intervals of event proposals are not exactly aligned with those of ground-truth proposals.
Specifically, for a sampled event sequence $\hat{\mathcal{E}}$, we find a reference event sequence $\tilde{\mathcal{E}}=\{\tilde{e}_1, \dots, \tilde{e}_{N_s}\}$ and its descriptions $\tilde{\mathcal{D}}=\{\tilde{d}_1, \dots, \tilde{d}_{N_s}\}$, where the reference event $\tilde{e}$ is given by one of the ground-truth proposals with the highest overlapping ratio with sampled event $\hat{e}$.
Then, the reward for the $n^{\text{th}}$ sampled description $\hat{d}_n$ is given by
\begin{align}
	\label{eq:rewards}
	\text{R}&(\hat{d}_n) = \\
	&\left[ f(\hat{d}_n, \tilde{d}_n) - f(\check{d}_n, \tilde{d}_n) \right]
	+ \left[f(\hat{\mathcal{D}}, \tilde{\mathcal{D}}) - f(\check{\mathcal{D}}, \tilde{\mathcal{D}}) \right], \nonumber
\end{align}
where $f(\cdot,\cdot)$ returns a similarity score between two captions or two set of captions, and $\check{\mathcal{D}}=\{\check{d}_1, \dots, \check{d}_{N_s}\}$ denote the generated descriptions from the reference event sequence.
Both terms in Eq.~\eqref{eq:rewards} encourage our model to increase the probability of sampled descriptions whose scores are higher than the results of generated captions from the ground-truth event proposals.
Note that the first and second terms are computed on the current event and episode, respectively.
We use two famous captioning metrics, METEOR and CIDEr, to define $f(\cdot, \cdot)$.

\section{Experiments}
\label{sec:experiments}

\subsection{Dataset}
\label{sec:dataset}

We evaluate the proposed algorithm on the ActivityNet Captions dataset~\cite{krishna2017dense}, which contains 20k YouTube videos with an average length of 120 seconds.
The dataset consists of 10,024, 4,926 and 5,044 videos for training, validation and test splits, respectively.
The videos have 3.65 temporally localized events and descriptions on average, where the average length of the descriptions is 13.48 words.

\subsection{Metrics}
\label{sec:metrics}

We use the performance evaluation tool\footnote{\url{https://github.com/ranjaykrishna/densevid\_eval}} provided by the 2018 ActivityNet Captions Challenge, which measures the capability to localize and describe events\footnote{On 11/02/2017, the official evaluation tool fixed a critical issue; only one out of multiple incorrect predictions for each video was counted.  This leads to performance overestimation of \cite{wang2018bidirectional, zhou2018end}. Thus, we received raw results from the authors and reported the scores measured by the new metric.}.
For evaluation, we measure recall and precision of event proposal detection, and METEOR, CIDEr and BLEU of dense video captioning.
The scores of the metrics are summarized via their averages based on tIoU thresholds of $0.3, 0.5, 0.7$ and $0.9$ given identified proposals and generated captions.
We use METEOR as the primary metric for comparison, since it is known to be more correlated to human judgments than others when only a small number of reference descriptions are available~\cite{vedantam2015cider}.

\subsection{Implementation Details}
\label{sec:implementation_details}

For EPN, we use a two-layer GRU with 512 dimensional hidden states and generate 128 proposals at each ending segment, which makes the dimensionality of $c_t$ in Eq.~\eqref{eq:sup_proposalnet} 128.
In our implementation, EPN based on SST takes a whole span of video for training as an input to the network, this allows the network to consider all ground-truth proposals, while the original SST~\cite{buch2017sst} is trained with densely sampled clips given by the sliding window method.

For ESGN, we adopt a single-layer GRU and a single-layer LSTM as $\text{EncoderRNN}$ and $\text{RNN}_{\text{ptr}}$, respectively, where the dimensions of hidden states are both 512.
We represent the location feature, denoted by $\text{Loc}(\cdot)$, of proposals with a 100 dimensional vector.
When learning SGN with reinforcement learning, we sample 100 event sequences for each video and generate one caption for each event in the event sequence with a greedy decoding.
In all experiments, we use Adam~\cite{kingma2014adam} to learn models with the mini-batch size 1 video and the learning rate 0.0005.

\subsection{Comparison with Other Methods}
\label{sec:comparison_others}
%
\begin{table}[!t]
	\centering
	\caption{
		Results on ActivityNet Captions evaluation server. 
	}
	\vspace{0.1cm}
	\scalebox{0.9}{
 	 	\setlength\tabcolsep{3pt} \hspace{-0.2cm}
		\begin{tabular}{c|ccc|c|c}
			\toprule
			& Audio & Flow & Visual & Ensemble & METEOR \\
			\hline\hline 
			RUC+CMU			& $\surd$ & $\surd$ & $\surd$ & yes & 8.53 \\
			YH Technologies &         & $\surd$ & $\surd$ & no & 8.13 \\
			Shandong Univ. 	&         & $\surd$ & $\surd$ & yes & 8.11 \\
			\hline
			SDVC (ours)		&         &         & $\surd$ & no & 8.19 \\
			\bottomrule
		\end{tabular}
	}
	\label{table:testset}
	\vspace{-0.25cm}
\end{table}

%
\begin{table*}[!t]
	\centering
	\caption{
		Ablation results of mean averaged recall, precision and METEOR over four tIoU thresholds of 0.3, 0.5, 0.7 and 0.9 on the ActivityNet Captions validation set.
		We also present the number of proposals in average.
		The bold-faced number means the best performance.
	}
	\vspace{0.1cm}
	\scalebox{0.9}{
		\begin{tabular}{l|ccccc|c|c|c|c}
			\toprule
			\multirow{2}{*}{Method} & \multicolumn{2}{c}{Proposal modules}    & \multicolumn{3}{|c|}{Captioning modules}  & Number of & \multirow{2}{*}{Recall} & \multirow{2}{*}{Precision} & \multirow{2}{*}{METEOR} \\
			& EPN & ESGN & \multicolumn{1}{|c}{eventRNN} & episodeRNN & RL & proposals &       &       &      \\
			\hline\hline 
			\emph{EPN-Ind} & $\surd$ & & \multicolumn{1}{|c}{$\surd$} & & & 77.99 & 84.97 & 28.10 & 4.58 \\ \hline
			\emph{ESGN-Ind} & & $\surd$ & \multicolumn{1}{|c}{$\surd$} & & & 2.85 & 55.58 & 57.57 & 6.73 \\
			\emph{ESGN-SCN} & & $\surd$ & \multicolumn{1}{|c}{$\surd$} & $\surd$ & & 2.85 & 55.58 & 57.57 &  6.92 \\
			\emph{ESGN-SCN-RL} (SDVC) & & $\surd$ & \multicolumn{1}{|c}{$\surd$} & $\surd$ & $\surd$ & 2.85 & 55.58 & 57.57 & \textbf{8.82} \\
			\bottomrule
		\end{tabular}
	}
	\label{table:ablation}
	\vspace{-0.2cm}
\end{table*}

We compare the proposed Streamlined Dense Video Captioning (SDVC) algorithm with several existing state-of-the-art methods including DCE~\cite{krishna2017dense}, DVC~\cite{li2018jointly}, Masked Transformer~\cite{zhou2018end} and TDA-CG~\cite{wang2018bidirectional}.
We additionally report the results of MFT~\cite{xiong2018move}, which is originally proposed for video paragraph generation but its event selection module is also able to generate an event sequence from the candidate event proposals; it makes a choice between selecting each proposal for caption generation and skipping it, and constructs an event sequence implicitly.
For MFT, we compare performances in both event detection and dense captioning.

Table~\ref{table:detection} presents the event detection performances of ESGN and MFT in ActivityNet Captions validation set.
ESGN outperforms the progressive event selection module in MFT on most tIoUs with large margins, especially in recall.
This validates the effectiveness of our proposed event sequence selection algorithm.

Table~\ref{table:captioning} illustrates performances of dense video captioning algorithms evaluated on ActivityNet Captions validation set.
We measure the scores with both ground-truth proposals and learned ones, where the number of the predicted proposals in individual algorithms may be different; DCE, DVC, Masked Transformer and TDA-CG uses 1,000, 1,000, 226.78 and 97.61 proposals in average, respectively, while the average number of proposals in SDVC is only 2.85.
According to Table~\ref{table:captioning}, SDVC improves the quality of captions significantly compared to all other methods.
Masked Transformer achieves comparable performance to ours using ground-truth proposals, but does not work well with learned proposals.
Note that it uses optical flow features in addition to visual features, while SDVC is only trained on visual features.
Since the motion information from optical flow features consistently improves the performances in other video understanding tasks~\cite{nguyen2018weakly,simonyan2014two}, incorporating motion information to our model may lead to additional performance gain.
MFT has the highest METEOR score among existing methods, which is partly because MFT considers temporal dependency across captions.

Table~\ref{table:testset} shows the test split results from the evaluation server.
SDVC achieves competitive performance based only on basic visual features while other methods exploit additional modalities (\eg, audio and optical flow) to represent videos and/or ensemble models to boost accuracy as described in \cite{ghanem2018activitynet}.
%
\subsection{Ablation Studies}
\label{sec:ablation}

We perform several ablation studies on ActivityNet Captions validation set to investigate the contributions of individual components in our algorithm.
In this experiment, we train the following four variants of our model:
1) \emph{EPN-Ind}: generating captions independently from all candidate event proposals, which is a baseline similar to most existing frameworks,
2) \emph{ESGN-Ind}: generating captions independently using eventRNN only from the events within the event sequence identified by our ESGN,
3) \emph{ESGN-SCN}: generating captions sequentially using our hierarchical RNN from the detected event sequence, and
4) \emph{ESGN-SCN-RL}: our full model (SDVC) that uses reinforcement learning to further optimize the captioning network.

Table~\ref{table:ablation} summarizes the results from this ablation study, and we have the following observations.
First, the approach based on ESGN (\emph{ESGN-Ind)} is more effective than the baseline that simply relies on all event proposals (\emph{EPN-Ind}).
Also, ESGN reduces the number of candidate proposals significantly, from  77.99 to 2.85 in average, with substantial increase in METEOR score, which indicates that ESGN successfully identifies event sequences from candidate event proposals.
Second, context modeling through hierarchical structure (\ie, $\text{event RNN}+\text{episode RNN}$) in a captioning network (\emph{ESGN-SCN}) enhances performance compared to the method with independent caption generation without considering context (\emph{ESGN-Ind}).
Finally, \emph{ESGN-SCN-RL} successfully integrates reinforcement learning to effectively improve the quality of generated captions.
%
\begin{table}[!t]
	\centering
	\caption{
		Performance comparison varying reward levels in reinforcement learning on the ActivityNet Captions dataset. 
	}
	\vspace{0.1cm}
	\scalebox{0.9}{
		\begin{tabular}{cc|c}
			\toprule
			Event-level reward & Episode-level reward & METEOR \\
			\hline\hline 
			$\surd$ & \multicolumn{1}{c|}{} & 8.73 \\
			& \multicolumn{1}{c|}{$\surd$} & 8.29 \\
			$\surd$ & \multicolumn{1}{c|}{$\surd$} & 8.82 \\
			\bottomrule
		\end{tabular}
	}
	\label{table:reward_type}
	\vspace{-0.25cm}
\end{table}
%
\begin{figure*}[!t]
	\centering
	\includegraphics[width=0.93\linewidth]{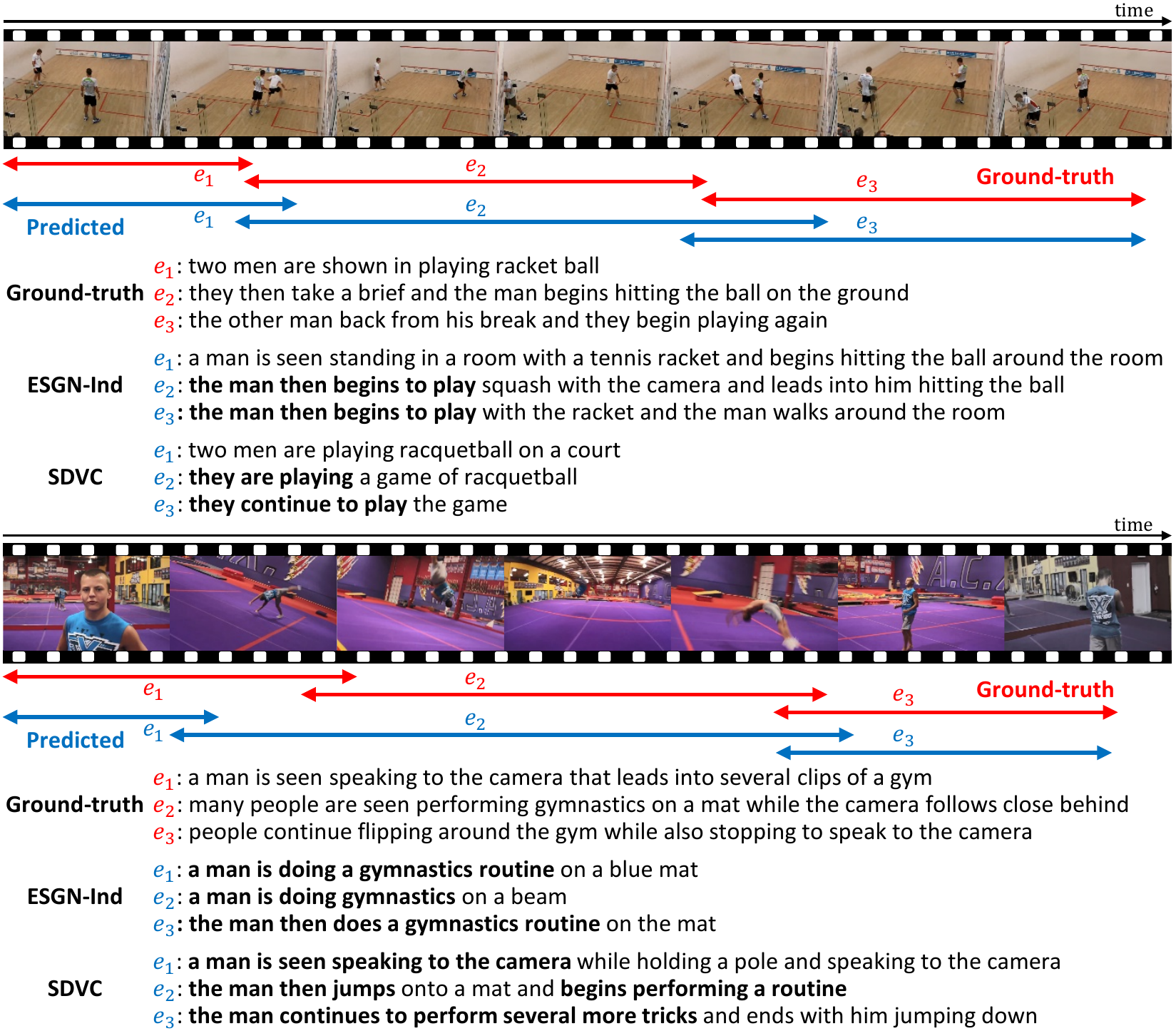}
	\vspace{0.05cm}
	\caption{
		Qualitative results on ActivityNet Captions dataset. The arrows represent ground-truth events (red) and events in the predicted event sequence from our event sequence generation network (blue) for input videos.
		Note that the events in the event sequence are selected in the order of its index.
		For the predicted events, we show the captions generated independently (ESGN-Ind) and sequentially (SDVC).
		More consistent captions are obtained by our sequential captioning network, where words for comparison are marked in bold-faced black.
	}
	\label{fig:qualitative_result}
	\vspace{-0.2cm}
\end{figure*}

We also analyze the impact of two reward levels---event and episode---used for reinforcement learning.
The results are presented in Table~\ref{table:reward_type}, which clearly demonstrates the effectiveness of training with rewards from both levels.

\subsection{Qualitative Results}
\label{sec:qualitative}

Fig.~\ref{fig:qualitative_result} illustrates qualitative results, where the detected event sequences and generated captions are presented together.
We compare the generated captions by our model (SDVC), which sequentially generates captions, with the model (ESGN-Ind) that generates descriptions independently from the detected event sequences.
Note that the proposed ESGN effectively identifies event sequences for input videos and our sequential caption generation strategy facilitates to describe events more coherently by exploiting both visual and linguistic contexts.
For instance, in the first example in Fig.~\ref{fig:qualitative_result}, SDVC captures the linguistic context (`two men' in $e_1$ is represented by `they' in both $e_2$ and $e_3$) as well as temporal dependency between events (an expression of `continue' in $e_3$), while ESGN-Ind just recognizes and describes $e_2$ and $e_3$ as independently occurring events.

\section{Conclusion}
\label{sec:conclusion}

We presented a novel framework for dense video captioning, which considers visual and linguistic contexts for coherent caption generation by modeling temporal dependency across events in a video explicitly.
Specifically, we introduced the event sequence generation network to detect a series of event proposals adaptively.
Given the detected event sequence, a sequence of captions is generated by conditioning on preceding events in our sequential captioning network.
We trained the captioning network in a supervised manner while further optimizing via reinforcement learning with two-level rewards for better context modeling.
Our algorithm achieved the state-of-the-art accuracy on the ActivityNet Captions dataset in terms of METEOR.

\vspace{-0.2cm}
\paragraph{Acknowledgments} \vspace{-0.2cm}
This work was partly supported by Snap Inc., Korean ICT R\&D program of the MSIP/IITP grant [2016-0-00563, 2017-0-01780], and SNU ASRI.

{\small
\bibliographystyle{ieee}
\bibliography{cvpr2019_sdvc}
}

\appendix
\section{Details of Event RNN}

As described in Section 3.4 of the main paper, the event RNN is in charge of generating a description given an event and a context information, and returning features for the generated captions.
This section discusses the caption generation process of the event RNN in detail.

Following \cite{wang2018bidirectional}, we provide two kinds of event information, $\text{C3D}(e)$ and $\text{Vis}(e)$, to the event RNN, where $\text{C3D}(e)$ is a set of segment-level feature descriptors in an interval of an event $e$, and $\text{Vis}(e)$ is a visual representation obtained from the event proposal network, SST.
We first set an initial hidden state of the event RNN to the context vector of episode $r$ given by the episode RNN.
Then, at each time step of the event RNN, we perform Temporal Dynamic Attention (TDA) to obtain an attentive segment-level feature from $\text{C3D}(e)$, followed by Context Gating (CG) to adaptively model relative contributions of the attentive segment-level feature and the visual feature and return a gated event feature.
Based on the gated event feature, the event RNN generates a word, and returns the hidden state as the caption feature $g$ when generating the $\text{END}$ token .

The whole caption generation process in the event RNN is summarized by the following sequence of operations:
\begin{align}
	h^{e}_{0} &= r, \\
	x_t &= W_{\text{wemb}}w_{t}, \\
	z_t &= \text{TDA}(\text{C3D}(e), \text{Vis}(e), h^{e}_{t-1}), \\
	o_t &= \text{CG}(z_t, \text{Vis}(e), x_t, h^e_{t-1}), \\
	h^e_t &= \text{LSTM}_{e}(o_{t}, x_{t}, h^e_{t-1}), \\
	p_t &= \text{Softmax}(W_p h^e_t),
\end{align}
where $W_{\text{wemb}}$ and $W_p$ are learnable parameters, $h^e$ means a hidden state of $\text{LSTM}_e$ in the event RNN, and $w_t$, $x_t$, $z_t$, $o_t$ and $p_t$ denote an input word, a word embedding vector, an attentive segment-level feature vector, a gated event feature vector and a probability distribution over vocabulary at time $t$, respectively.
At time step $t$, given an event with $S$ segments, TDA computes the attentive vector $z_t$ by
\begin{align}
	\alpha^{t}_{s} = W_\alpha \text{tanh} (W_{\text{c}}& \text{C3D}(e_s) + W_{\text{v}} \text{Vis}(e) + W_{h} h^e_{t-1}), \\
	a^t_s &= \frac{\exp(\alpha^t_s)}{\sum^{S}_{s=1}\exp{(\alpha^t_s)}}, \\
	z_t &= \sum^{S}_{s=1} a^t_s \text{C3D}(e_s), 
\end{align}
where $W_{\alpha}$, $W_c$, $W_v$ and $w_h$ are learnable parameters, and $e_s$ indicates the $s^\text{th}$ segment in event $e$.
Once obtaining the attentive segment-level feature, CG computes the gating vector $k_t$ and the gated event vector $o_t$ as follows:
\begin{align}
	\bar{z}_t &= \text{tanh} (W_{\text{z}} z_t), \\
	\bar{v} &= \text{tanh} (W_{\bar{\text{v}}} \text{Vis}(e)), \\
	k_t &= \sigma (W_k [\bar{z}; \bar{v}; x_t; h^e_{t-1}]), \\
	o_t &= [(1-k_t) \odot \bar{z}_t; k_t \odot \bar{v}], 
\end{align}
where $W_z$, $W_{\bar{v}}$ and $W_k$ are learnable parameters, $\sigma$ is a sigmoid function, $[\cdot;\cdot]$ denotes vector concatenation and $\odot$ means element-wise multiplication.

\begin{figure}[!t]
	\centering
	\includegraphics[width=0.49\linewidth]{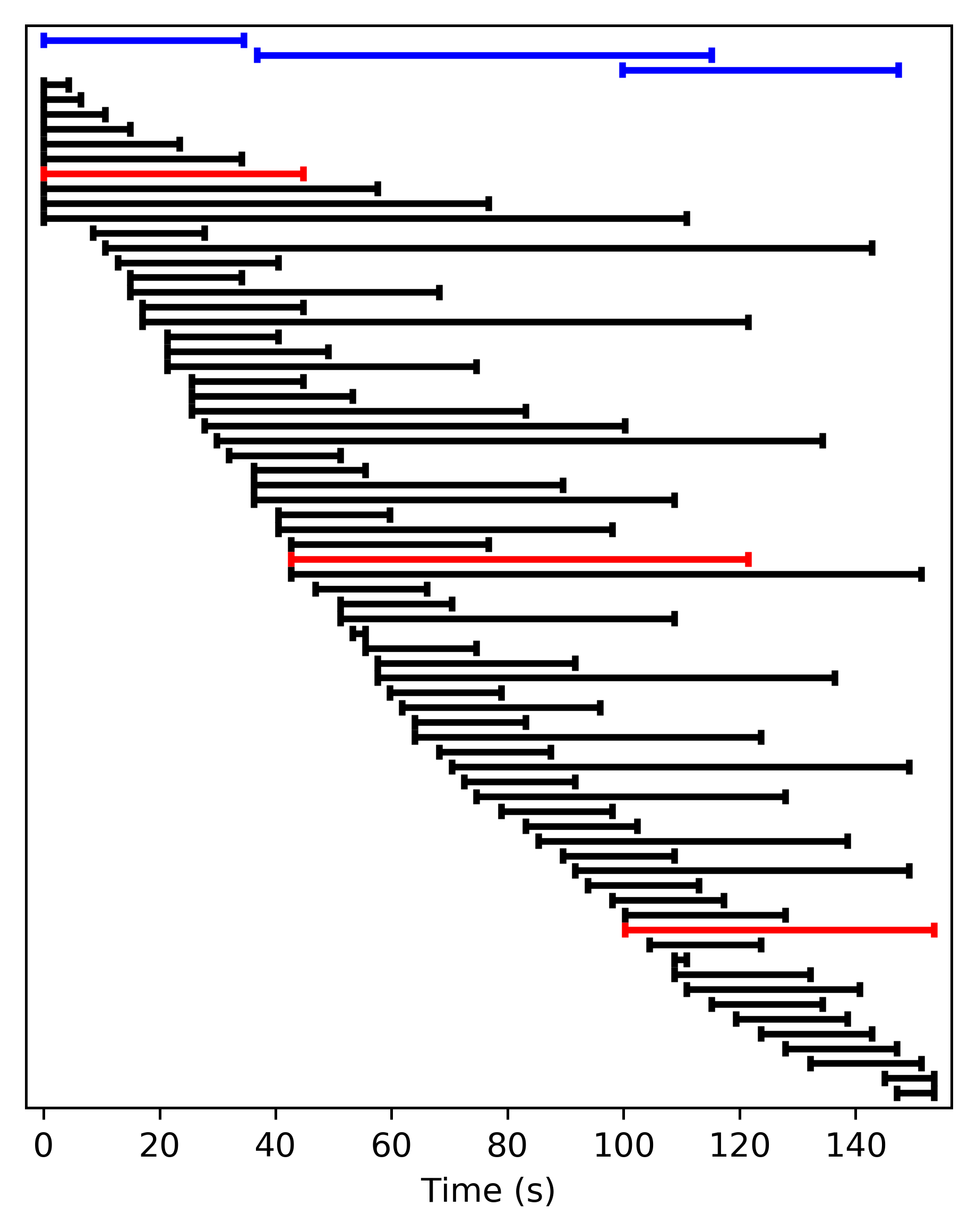}
	\includegraphics[width=0.49\linewidth]{event_selection1.png}
	\caption{
		Examples of the selected event proposals ({\color{red}\textbf{red}}) out of the candidates (\textbf{black}) in the proposed event sequence generation network and the ground-truth events ({\color{blue}\textbf{blue}}).
	}
	\label{fig:event_selection}
	\vspace{-0.2cm}
\end{figure}

\section{Visualization of Event Selection}
Fig.~\ref{fig:event_selection} illustrates the event selection results.
Our proposed event sequence generation network successfully identifies the event proposals out of the candidates, which highly overlap with the ground-truths.

\end{document}